\documentclass[letterpaper, 10 pt, conference]{ieeeconf}  

\IEEEoverridecommandlockouts                              

\usepackage{amsmath} 
\usepackage{amssymb}  
\usepackage{mathtools}
\usepackage{commath}

\usepackage{graphicx}
\usepackage[tight,footnotesize]{subfigure}
\usepackage{hyphenat}
\usepackage[table]{xcolor}
\usepackage{multirow}
\usepackage{url}
\usepackage{booktabs}
\urldef{\mailsa}\path| srahman@email.sc.edu|
\urldef{\mailsb}\path|{albertoq, yiannisr}@cse.sc.edu|
\usepackage{tikz}
\usepackage{hyperref}

\newcommand\eq[1]{Eq.~\eqref{#1}}
\newcommand\fig[1]{Fig.~\ref{#1}}

\long\def\invis#1{}

\usepackage{xspace}

\newcommand\sect[1]{Section~\ref{#1}}
\newcommand\tab[1]{Table~\ref{#1}}

\makeatletter
\DeclareRobustCommand\onedot{\futurelet\@let@token\@onedot}
\def\@onedot{\ifx\@let@token.\else.\null\fi\xspace}

\def\etal{\emph{et al}\onedot}
\makeatother

\usepackage{siunitx}

\newcommand\copyrighttext{%
  \footnotesize \textcopyright 2019 IEEE. Personal use of this material is permitted.
  Permission from IEEE must be obtained for all other uses, in any current or future
  media, including reprinting/republishing this material for advertising or promotional
  purposes, creating new collective works, for resale or redistribution to servers or
  lists, or reuse of any copyrighted component of this work in other works.
  DOI: \href{https://doi.org/10.1109/IROS40897.2019.8967703}{10.1109/IROS40897.2019.8967703}}
\newcommand\copyrightnotice{%
\begin{tikzpicture}[remember picture,overlay]
\node[anchor=south,yshift=10pt] at (current page.south) {\fbox{\parbox{\dimexpr\textwidth-\fboxsep-\fboxrule\relax}{\copyrighttext}}};
\end{tikzpicture}%
}

\title{\LARGE \bf
SVIn2: An Underwater SLAM System using Sonar, Visual, Inertial, and Depth Sensor
}

\author{Sharmin Rahman$^1$, Alberto Quattrini Li$^2$, and Ioannis Rekleitis$^1$
\thanks{$^1$S. Rahman and I. Rekleitis are with the Computer Science and Engineering Department, University of South Carolina, Columbia, SC, USA,
{\tt\small srahman@email.sc.edu, yiannisr@cse.sc.edu}}%
\thanks{$^2$A. Quattrini Li is with the Department of Computer Science, Dartmouth College, Hanover, NH, USA, {\tt\small alberto.quattrini.li@dartmouth.edu}}%
\thanks{The authors would like to thank the National Science Foundation for its support (NSF 1513203, 1637876). The authors would also like to acknowledge the help of the Woodville Karst Plain Project (WKPP).}}

\begin{document}

\maketitle
\copyrightnotice
\thispagestyle{empty}
\pagestyle{empty}

\begin{abstract}
This paper presents a novel tightly\hyp coupled keyframe\hyp based  Simultaneous Localization and Mapping (SLAM) system with loop\hyp closing and relocalization capabilities targeted for the underwater domain. 

Our previous work, SVIn, augmented the state\hyp of\hyp the\hyp art visual\hyp inertial state estimation package OKVIS to accommodate acoustic data from sonar in a non\hyp linear optimization\hyp based framework. 
This paper addresses drift and loss of localization -- one of the main problems affecting other packages in underwater domain -- by providing the following main contributions: a robust initialization method to refine scale using depth measurements, a fast preprocessing step to enhance the image quality, and a real\hyp time loop\hyp closing and relocalization method using bag of words. An additional contribution is the introduction of depth measurements from a pressure sensor to the tightly\hyp coupled optimization formulation. Experimental results on datasets collected with a custom-made underwater sensor suite and an autonomous underwater vehicle from challenging underwater environments with poor visibility demonstrate performance never achieved before in terms of accuracy and robustness.
\end{abstract}

\section{INTRODUCTION}
Exploring and mapping underwater environments such as caves, bridges, dams, and shipwrecks, are extremely important tasks for the economy, conservation, and scientific discovery~\cite{ballard2014smithsonian}. Currently, most of the efforts are performed by divers that need to take measurements manually using a grid and measuring tape, or using hand-held sensors~\cite{henderson2013mapping}, and data is post\hyp processed afterwards. Autonomous Underwater Vehicles (AUVs) present unique opportunities to automate this process; however, there are several open problems that still need to be addressed for reliable deployments, including robust Simultaneous Localization and Mapping (SLAM), the focus of this paper.

\begin{figure}
{\includegraphics[width=0.45\textwidth]{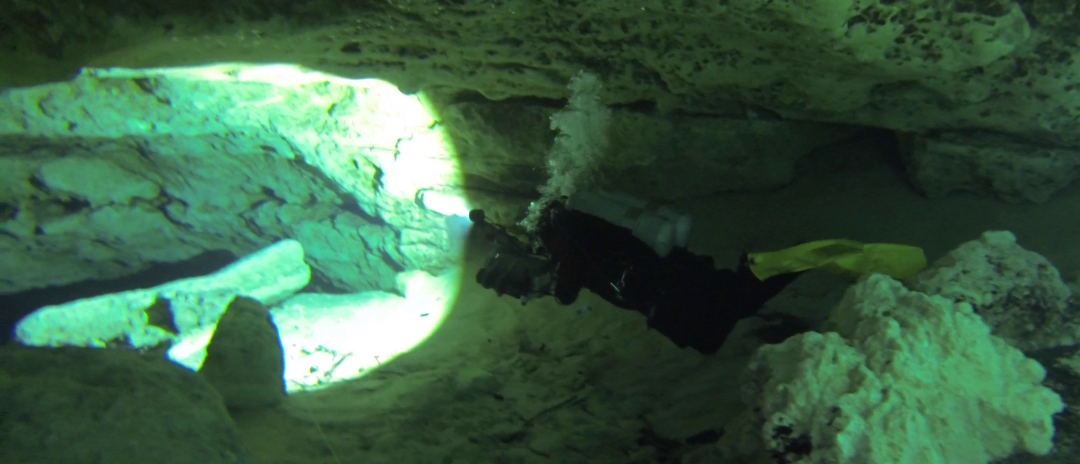}}
\caption{Underwater cave in Ginnie Springs, FL, where data have been collected using an underwater stereo rig.\vspace{-0.2in}}
\label{fig:beauty}
\end{figure}

Most of the underwater navigation algorithms~\cite{leonard2012directed, lee2005underwater, snyder2010doppler, johannsson2010imaging, rigby2006towards} are based on acoustic sensors, such as Doppler velocity log (DVL), Ultra-short Baseline (USBL), and sonar. However, data collection with these sensors is expensive and sometimes not suitable due to the highly unstructured underwater environments.
In recent years, many vision\hyp based state estimation algorithms have been developed using monocular, stereo, or multi-camera system mostly for indoor and outdoor environments. Vision is often combined with Inertial Measurement Unit (IMU) for improved estimation of pose in challenging environments, termed as \emph{Visual-Inertial Odometry} (VIO)~\cite{mur2017visual,okvis,qin2018vins,mourikis2007multi,msckf-kumar-ral}. 
However, the underwater environment -- e.g., see \fig{fig:beauty} -- presents unique challenges to vision\hyp based state estimation. As shown in a previous study~\cite{RekleitisISERVO2016},  it is not straightforward to deploy the available vision\hyp  based state estimation packages underwater. 
In particular, suspended particulates, blurriness, and light and color attenuation result in features that are not as clearly defined as above water. Consequently results from different vision\hyp based state estimation packages show a significant number of outliers resulting in inaccurate estimate or even complete tracking loss.

In this paper, we propose \emph{SVIn2}, a novel SLAM system specifically targeted for underwater environments -- e.g., wrecks and underwater caves -- and easily adaptable for different sensor configuration: acoustic (mechanical scanning profiling sonar), visual (stereo camera), inertial (linear accelerations and angular velocities), and depth data. This makes our system versatile and applicable on-board of different sensor suites and underwater vehicles.

In our recent work, {\em SVIn}~\cite{rahman2018icra}, acoustic, visual, and inertial data is fused together to map different underwater structures by augmenting the visual-inertial state estimation package OKVIS~\cite{okvis}. This improves the trajectory estimate especially when there is varying visibility underwater, as sonar provides robust information about the presence of obstacles with accurate scale. However, in long trajectories, drifts could accumulate resulting in an erroneous trajectory. 

{\em SVIn}~\cite{rahman2018icra} is extended by including an image enhancement technique targeted to the underwater domain, introducing depth measurements in the optimization process, loop\hyp closure capabilities, and a more robust initialization. These additions enable the proposed approach to robustly and accurately estimate the sensor's trajectory, where every other approach has shown incorrect trajectories or loss of localization.

To validate the proposed approach, first, we assess the performance of the proposed loop-closing method, by comparing it to other state-of-the-art systems on the EuRoC micro\hyp aerial vehicle public dataset~\cite{Burri25012016}, disabling the fusion of sonar and depth measurements in our system. 
Second, we test the proposed full system on several  underwater datasets in a diverse set of conditions. More specifically, underwater data -- consisting of visual, inertial, depth, and acoustic measurements -- has been collected using a custom made sensor suite~\cite{RahmanOceans2018} from different locales; furthermore, data collected by an Aqua2 underwater vehicle~\cite{Rekleitis2005d} include visual, inertial, and depth measurements. 
The results on the underwater datasets illustrate the loss of tracking and/or failure to maintain consistent scale for other state-of-the-art systems while our proposed method maintains correct scale without diverging; for a comprehensive comparison please refer to Joshi \etal~\cite{joshi2019iros}. 

The paper is structured as follows. The next section discusses related work. \sect{sec:method} presents the mathematical formulation of the proposed system and describes the approach developed for image preprocessing, pose initialization, loop\hyp closure, and relocalization. \sect{sec:results} presents results from a publicly available aerial dataset and a diverse set of challenging underwater environments. We conclude this paper with a discussion on lessons learned and directions of future work.

\section{RELATED WORK}
Sonar based underwater SLAM and navigation systems have been exploited for many years. Folkesson et al.~\cite{folkesson2007feature} used a blazed array sonar for real-time feature tracking. A feature reacquisition system with a low-cost sonar and navigation sensors was described in~\cite{fallon2013relocating}. More recently, Sunfish~\cite{richmond2018sunfish} -- an underwater SLAM system using a multibeam sonar, an underwater dead-reckoning system based on a fiber-optic gyroscope (FOG) IMU, acoustic DVL, and pressure-depth sensors -- has been developed for autonomous cave exploration. Vision and visual-inertial based SLAM systems also developed in~\cite{salvi2008visual, beall2011bundle, shkurti2011state} for underwater reconstruction and navigation. Corke et al.~\cite{corke2007experiments} compared acoustic and visual methods for underwater localization showing the viability of using visual methods underwater in some scenarios.

The literature presents many vision\hyp based state estimation techniques, which use either \emph{monocular} or \emph{stereo} cameras and that are \emph{indirect} (feature-based) or \emph{direct} methods, including, for example, MonoSLAM~\cite{Civera2010}, PTAM~\cite{4538852}, ORB-SLAM~\cite{ORBSLAM2015}, LSD-SLAM~\cite{raey}, and DSO~\cite{engel2018direct}. In the following, we highlight some of the state estimation systems which use visual-inertial measurements and feature-based method.

To improve the pose estimate, vision\hyp based state estimation techniques have been augmented with IMU sensors, whose data is fused together with visual information. A class of approaches is based on the \emph{Kalman Filter}, e.g., Multi-State Constraint Kalman Filter (MSCKF) \cite{mourikis2007multi} and its stereo extension \cite{msckf-kumar-ral}; ROVIO~\cite{rovio}; REBiVO~\cite{rebivo}.
The other spectrum of methods optimizes the sensor states, possibly within a window, formulating the problem as a \emph{graph optimization} problem. For feature-based visual-inertial systems, as in OKVIS~\cite{okvis} and Visual-Inertial ORB-SLAM~\cite{mur2017visual}, the optimization function includes the IMU error term and the reprojection error. The \emph{frontend} tracking mechanism maintains a local map of features in a marginalization window which are never used again once out of the window. VINS-Mono~\cite{qin2018vins} uses a similar approach and maintains a minimum number of features for each image and existing features are tracked by Kanade\hyp Lucas\hyp Tomasi (KLT) sparse optical flow algorithm in local window. Delmerico and Scaramuzza \cite{Delmerico:254865} did a comprehensive comparison specifically monitoring resource usage by the different methods. While KLT sparse features allow VINS-Mono running in real-time on low-cost embedded systems, often results into tracking failure in challenging environments, e.g., underwater environments with low visibility. In addition, for loop detection additional features and their descriptors are needed to be computed for keyframes. An evaluation of features for the underwater domain was presented in Shkurti \etal~\cite{ShkurtiCRV2011} and in Quattrini Li \etal~\cite{QuattriniLiOceans2016}.

\emph{Loop closure} -- the capability of recognizing a place that was seen before -- is an important component to mitigate the drift of the state estimate.
FAB-MAP~\cite{cummins2008fab,cummins2011appearance} is an appearance-based method to recognize places in a probabilistic framework.
ORB-SLAM~\cite{ORBSLAM2015} and its extension with IMU~\cite{mur2017visual} use bag\hyp of\hyp words (BoW) for loop closure and relocalization. VINS-Mono also uses a BoW approach.

Note that all visual\hyp inertial state estimation systems require proper \emph{initialization}.
VINS-Mono uses a loosely-coupled sensor fusion method to align monocular vision with inertial measurement for estimator initialization.
ORB-SLAM with IMU \cite{mur2017visual} performs initialization by first running a monocular SLAM to observe the pose first and then, IMU biases are estimated.

Given the modularity of OKVIS for adding new sensors and robustness in tracking in underwater environment -- as demonstrated by fusing sonar data with Visual\hyp Inertial Odometry in Rahman \etal~ \cite{rahman2018icra} -- we extend OKVIS to include also depth estimate, loop closure capabilities, and a more robust initialization to specifically target underwater environments.

\section{Proposed Method}
\label{sec:method}

\begin{figure}
 \includegraphics[width=\columnwidth]{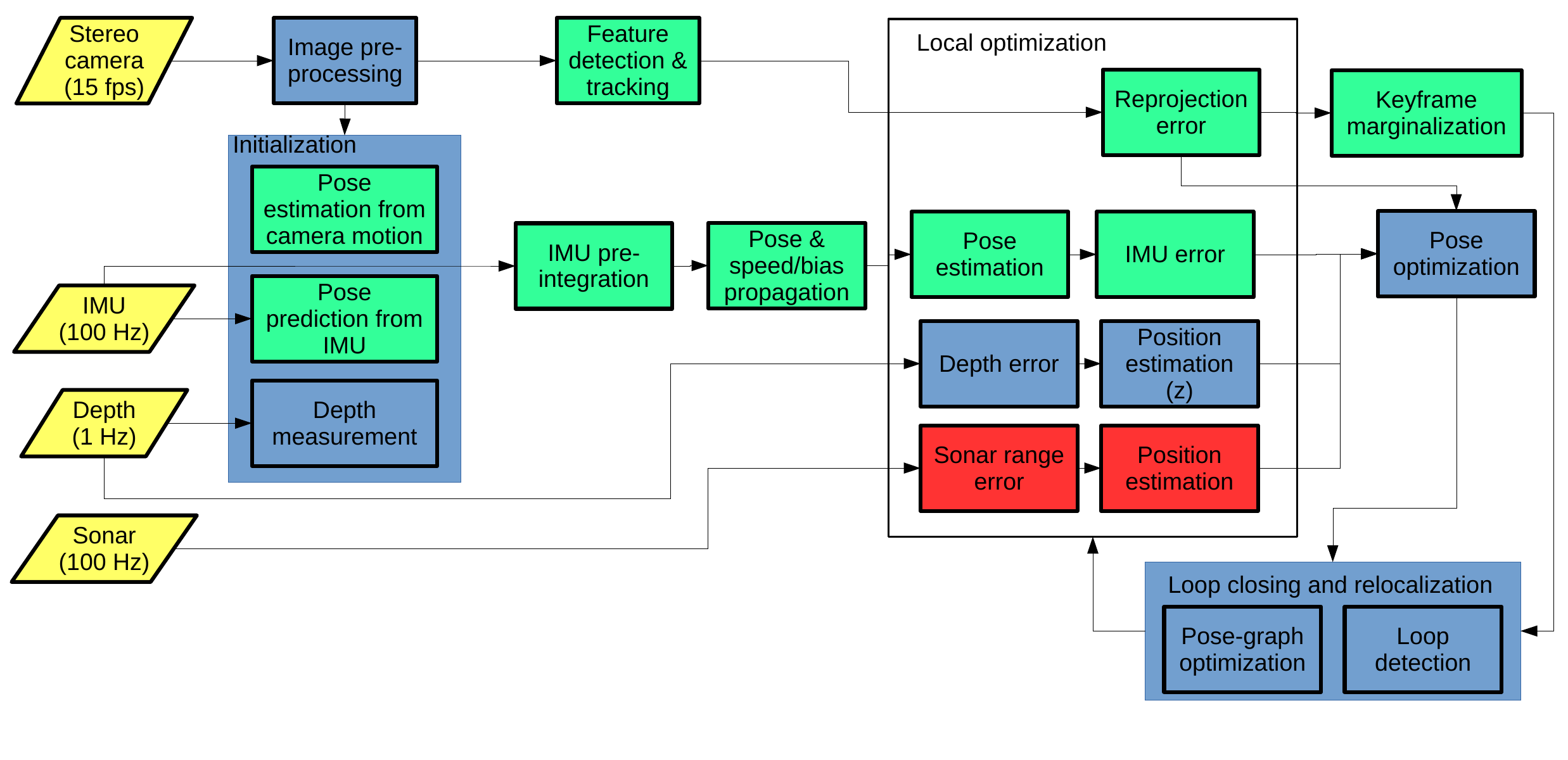}
 \vspace{-0.2in}\caption{Block diagram of the proposed system, SVIn2; in yellow the sensor input with frequency from the custom-made sensor suite, in green the components from OKVIS, in red the contribution from our previous work~\cite{rahman2018icra}, and in blue the new contributions in this paper. \vspace{-0.2in} \label{fig:pipeline}}
\end{figure}

This section describes the proposed system, SVIn2, depicted in \fig{fig:pipeline}. The full proposed state estimation system can operate with a robot that has stereo camera, IMU, sonar, and depth sensor -- the last two can be also disabled to operate as a visual-inertial system.

Due to low visibility and dynamic obstacles, it is hard to find good features to track. In addition to the underwater vision constraints,
e.g., light and color attenuation, vision-based systems also suffer from poor contrast. Hence, we augment the pipeline by adding an image preprocessing step, where \textit{contrast adjustment} along with \textit{histogram equalization} is applied to improve feature detection underwater. In particular, we use a \textit{Contrast Limited Adaptive Histogram Equalization} (CLAHE) filter~\cite{pizer1987adaptive} in the \textit{image pre-processing} step.

 In the following, after defining the state, we describe the proposed initialization, sensor fusion optimization, loop closure and relocalization steps.

\subsection{Notations and States}

The full sensor suite is composed of the following coordinate frames: Camera (stereo), IMU, Sonar (acoustic), Depth, and World which are denoted as $C$, $I$, $S$, $D$, and $W$ respectively. 
The transformation between two arbitrary coordinate frames $X$ and $Y$ is represented by a homogeneous transformation matrix $_{X}\textbf{T}_{Y}=[_{X}\textbf{R}_{Y} | _{X}\textbf{p}_{Y}]$ where $_{X}\textbf{R}_{Y}$ is rotation matrix with corresponding quaternion $_{X}\textbf{q}_{Y}$ and $_{X}\textbf{p}_{Y}$ is position vector.

Let us now define the robot $R$ state $\textbf{x}_{R}$ that the system is estimating as:
\begin{eqnarray} \label{eq:robot-state}
  \textbf{x}_{R} &=& [_{W}\textbf{p}_{I}^{T}, _{W}\textbf{q}_{I}^{T}, _{W}\textbf{v}_{I}^{T}, {\textbf{b}_g}^T, {\textbf{b}_a}^T]^T
\label{eq1}
\end{eqnarray}
\noindent which contains the position $_{W}\textbf{p}_{I}$, the attitude represented by the quaternion $_{W}\textbf{q}_{I}$, the linear velocity $_W\textbf{v}_{I}$, all expressed as the IMU reference frame $I$ with respect to the world coordinate $W$; moreover, the state vector contains the gyroscopes and accelerometers bias $\textbf{b}_g$ and $\textbf{b}_a$. 

The associated error-state vector is defined in minimal coordinates, while the perturbation takes place in the tangent space of the state manifold. The transformation from minimal coordinates to tangent space can be done using a bijective mapping~\cite{okvis, forster2017manifold}:
\begin{eqnarray} \label{eq:tangent}
  \delta\boldsymbol{\chi}_{R} &=& [\delta\textbf{p}^T, \delta\boldsymbol{\alpha}^T, \delta\textbf{v}^T, \delta{\textbf{b}_g}^T, \delta{\textbf{b}_a}^T]^T
  \label{eq2}
\end{eqnarray}

\noindent which represents the error for each component of the state vector with $\delta\boldsymbol{\alpha} \in \mathbb{R}^3$ being the minimal perturbation for rotation.

\subsection{Tightly-coupled Non-Linear Optimization with Sonar-Visual-Inertial-Depth measurements}\label{sec:opt}

For the tightly-coupled non-linear optimization, we use the following cost function $\textit{J}(\textbf{x})$, which includes the reprojection error $\textbf{e}_{r}$ and the IMU error $\textbf{e}_{s}$ with the addition of the sonar error $\textbf{e}_{t}$ (see \cite{rahman2018icra}), and the depth error ${e}_{u}$:

\begin{eqnarray}
\textit{J}(\textbf{x}) &=&\sum_{i=1}^{2}\sum_{k=1}^{K}\sum_{j \in \mathcal{J}(i,k)}  {\textbf{e}_r^{i,j,k^{T}}}\textbf{P}_r^k{\textbf{e}_r^{i,j,k}}  +\sum_{k=1}^{K-1}{\textbf{e}_s^{{k}^{T}}}\textbf{P}_s^k{\textbf{e}_s^k}\nonumber\\
&+& \sum_{k=1}^{K-1}{\textbf{e}_t^{{k}^{T}}}\textbf{P}_t^k{\textbf{e}_t^k} + \sum_{k=1}^{K-1}{{e}_u^{{k}^{T}}}{P}_u^k{{e}_u^k}
\label{eq7}
\end{eqnarray}

\noindent where $\textit{i}$ denotes the camera index -- i.e., left ($i=1$) or right ($i=2$) camera in a stereo camera system with landmark index $\textit{j}$ observed in the $\textit{k}$\textsuperscript{th} camera frame.  $\textbf{P}_r^k$, $\textbf{P}_s^k$, $\textbf{P}_t^k$, and ${P}_u^k$  represent the information matrix of visual landmarks, IMU, sonar range, and depth measurement for the $\textit{k}$\textsuperscript{th} frame respectively.

For completeness, we briefly discuss each error term -- see \cite{okvis} and \cite{rahman2018icra} for more details.
The reprojection error describes the difference between a keypoint measurement in camera coordinate frame $C$ and the corresponding landmark projection according to the stereo projection model. The IMU error term combines all accelerometer and gyroscope measurements by \emph{IMU pre-integration}~\cite{forster2017manifold} between successive camera measurements and represents the \emph{pose}, \emph{speed and bias} error between the prediction based on previous and current states. Both  reprojection error and IMU error term follow the formulation by Leutenegger \etal~\cite{okvis}.

The concept behind calculating the sonar range error, introduced in our previous work \cite{rahman2018icra}, is that, if the sonar detects any obstacle at some distance, it is more likely that the visual features would be located on the surface of that obstacle, and thus will be approximately at the same distance. The step involves computing a visual patch detected in close proximity of each sonar point to introduce an extra constraint, using the distance of the sonar point to the patch. Here, we assume that the visual-feature based patch is small enough and approximately coplanar with the sonar point.
As such, given the sonar measurement $\textbf{z}_t^k$, the error term $\textbf{e}_t^k(_W \textbf{p}_I^k, \textbf{z}_t^k)$ is based on the difference between those two distances which is used to correct the position $_W \textbf{p}_I^k$. We assume an approximate normal conditional probability density function $\textit{f}$ with zero \emph{mean} and $\textbf{W}_t^k$ \emph{variance}, and the conditional covariance $\textbf{Q}({\delta \hat{\textbf{p}^k}}|\textbf{z}_t^k)$, updated iteratively as new sensor measurements are integrated:

\begin{equation}
  \textit{f}(\textbf{e}_t^k|_W \textbf{p}_I^k) \approx \mathcal{N}(\textbf{0}, \textbf{W}_t^k)
\end{equation} 
The information matrix  is:
\begin{equation}
  \textbf{P}_t^k = {\textbf{W}_t^k}^{-1} = \left({{\frac{\partial \textbf{e}_t^k}{\partial {\delta \hat{\textbf{p}^k}}}} \textbf{Q}({\delta \hat{\textbf{p}^k}}|\textbf{z}_t^k){{\frac{\partial \textbf{e}_t^k}{\partial {\delta \hat{\textbf{p}^k}}}} }^T}\right)^{-1}
\end{equation}
The Jacobian can be derived by differentiating the expected \emph{range} $r$ measurement with respect to the robot position:

\begin{equation}
\resizebox{.85\hsize}{!}{$
  \frac{\partial \textbf{e}_t^k}{\partial {\delta \hat{\textbf{p}^k}}} = \left[\frac{-\textit{l} _x +{}_W{p_x}}{r}, \frac{-\textit{l}_y +{}_W{p_y}}{r}, \frac{-\textit{l}_z +{}_W{p_z}}{r}\right]
  $}
\end{equation}

\noindent where $_W\textit{\textbf{l}} = [\textit{l}_x, \textit{l}_y, \textit{l}_z, 1 ]$ represents the sonar landmark in homogeneous coordinate and can be calculated by a simple geometric transformation in world coordinates given \emph{range} $r$ and \emph{head-position} $\theta$ from the sonar measurements:
\begin{equation}
  _W\textit{\textbf{l}} =  ({_W{\textbf{T}}_{I}} {_I{\textbf{T}}_{S}} {{[{\textbf{I}}_3 |  r\cos(\theta), r\sin(\theta), 0]}}_{S}^{T})
\end{equation}

The pressure sensor, introduced in this paper, provides accurate depth measurements based on water pressure. Depth values are extracted along the \textit{gravity} direction which is aligned with the \textit{z} of the world $W$ -- observable due to the tightly coupled IMU integration. The depth data at time $k$ is given by\footnote{More precisely, $_W{{p_z}_D}^k = (d^k - d^0)+\textit{init\_disp\_from\_IMU}$ to account for the initial displacement  along $z$ axis from IMU, which is the main reference frame used by visual SLAM to track the sensor suite/robot.}:

\begin{equation}
    _W{{p_z}_D}^k = d^k - d^0
\end{equation}

With depth measurement ${z}_u^k$, the depth error term ${e}_u^k( _W{{p_z}_I}^k, {z}_u^k)$ can be calculated as the difference between the robot position along the $z$ direction and the depth data to  correct the position of the robot. The error term can be defined as:
\begin{equation}
{e}_u^k( _W{{p_z}_I}^k, {z}_u^k) = |_{W}{p_z}_{I}^{k} - _{W}{p_z}_{D}^{k}|
\end{equation}

The information matrix calculation follows a similar approach as the sonar and the Jacobian is straight-forward to derive.

All the error terms are added in the {\em Ceres Solver} nonlinear optimization framework~\cite{ceres} to formulate error-state (\eq{eq:tangent}) and estimate the robot state (\eq{eq:robot-state}).

\subsection{Initialization: Two-step Scale Refinement}
\begin{figure}
\begin{center}
 \includegraphics[width=0.8\columnwidth]{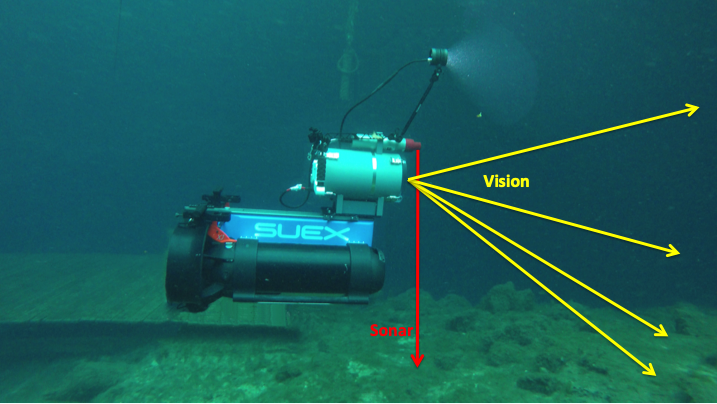}
 \vspace{-0.1in}\caption{Custom made sensor suite mounted on a dual DPV. Sonar scans around the sensor while the cameras see in front.\vspace{-0.3in} \label{fig:rigdpv}}
\end{center}
\end{figure}

A robust and accurate initialization is required for the success of tightly-coupled non-linear systems, as described in~\cite{mur2017visual} and~\cite{qin2018vins}. For underwater deployments, this becomes even more important as vision is often occluded as well as is negatively affected by the lack of features for tracking. 
Indeed, from our comparative study of visual-inertial based state estimation systems~\cite{joshi2019iros},  in underwater datasets, most of the state-of-the-art systems either fail to initialize or  make wrong initialization  resulting into divergence. 
Hence, we propose a robust initialization method using the sensory information from stereo camera, IMU, and depth for underwater state estimation. 
The reason behind using all these three sensors is to introduce constraints on \textit{scale} to have a more accurate estimation on initialization. Note that no acoustic measurements have been used because the sonar range and visual features contain a temporal difference, which would not allow to have any match between acoustic and visual features, if the robot is not moving. This is due to the fact that the sonar scans on a plane over \ang{360} \textit{around the robot} and camera detects features \textit{in front of the robot} \cite{rahman2018icra}; see \fig{fig:rigdpv}.

The proposed initialization works as follows. First, we make sure that the system only initializes when a minimum number of visual features are present  to track (in our experiments $15$ worked well). Second, the two-step refinement of the initial scale from the stereo vision takes place. 

The depth sensor provides accurate depth measurements which are used to refine the initial scale factor from stereo camera. Including a scale factor $s_1$, the transformation between camera $C$ and depth sensor $D$ can be expressed as

\begin{equation}\label{eq:depth-position}
    _W{p_z}_D = {s_1}*{_W{p_z}_C} + {_W{\textbf{R}_z}_C}{_C{\textbf{p}}_D}
\end{equation}

For \textit{keyframe} $k$, solving \eq{eq:depth-position} for $s_1$, provides the first refinement $r_1$ of the initial stereo scale ${_W{\textbf{p}_{r1}}_C}$, i.e.,

\begin{equation}
    {_W{\textbf{p}_{r1}}_C} = {s_1}*{_W{\textbf{p}}_C}
\label{eq_scale1}
\end{equation}

In the second step, the refined measurement from stereo camera in \eq{eq_scale1} is aligned with the IMU pre-integral values. Similarly, the transformation between camera $C$ and IMU $I$ with scale factor $s_2$ can be expressed as:

\begin{eqnarray} 
    _W{\textbf{p}}_I = {s_2}*{_W{\textbf{p}_{r1}}_C} + {_W{\textbf{R}}_C}{_C{\textbf{p}}_I}
\label{eq3}
\end{eqnarray} 

In addition to refining the scale, we also approximate initial \textit{velocity} and \textit{gravity} vector similar to the method described in~\cite{qin2018vins}. The state prediction from IMU integration $\textbf{\^{x}}_{R}^{i+1}(\textbf{x}_{R}^{i}, \textbf{z}_{I}^{i})$ with IMU measurements $\textbf{z}_{I}^{i}$ 
in OKVIS~\cite{okvis} with conditional covariance $\textbf{Q}(\delta\textbf{\^{x}}_{R}^{i+1} | \textbf{x}_{R}^{i}, \textbf{z}_{I}^{i})$ can be written as (the details about IMU pre-integration can be found in~\cite{forster2017manifold}): 
\begin{eqnarray} 
_{W}\textbf{\^{p}}_{I}^{i+1} &=& _{W}\textbf{p}_{I}^{i} + _{W}\textbf{v}_{I}^{i}\Delta{t}_i + \frac{1}{2}{ _W{\textbf{g}}}{\Delta{t}_{i}}^2 + _{W}\textbf{R}_{I}^{i}\boldsymbol{\alpha}_{{I}_{i}}^{{i+1}} \nonumber\\
_{W}\textbf{\^{v}}_{I}^{i+1} &=& _{W}\textbf{v}_{I}^{i} + _W{\textbf{g}}{\Delta{t}_{i}} + _{W}\textbf{R}_{I}^{i}\boldsymbol{\beta}_{{I}_{i}}^{{i+1}}\nonumber\\
_{W}\textbf{\^{q}}_{I}^{i+1} &=& \boldsymbol{\gamma}_{{I}_{i}}^{{i+1}}
\label{eq4}
\end{eqnarray}
\noindent where $\boldsymbol{\alpha}_{{I}_{i}}^{{i+1}}$, $\boldsymbol{\beta}_{{I}_{i}}^{{i+1}}$, and $\boldsymbol{\gamma}_{{I}_{i}}^{{i+1}}$ are IMU pre-integration terms defining the motion between two consecutive keyframes $\textit{i}$ and $\textit{i+1}$  in time interval $\Delta t_i$ and can be obtained only from the IMU measurements. \eq{eq4} can be re-arranged with respect to $\boldsymbol{\alpha}_{{I}_{i}}^{{i+1}}$, $\boldsymbol{\beta}_{{I}_{i}}^{{i+1}}$ as follows:

\begin{eqnarray} 
\boldsymbol{\alpha}_{{I}_{i}}^{{i+1}} &=& _{I}\textbf{R}_{W}^{i} (_{W}\textbf{\^{p}}_{I}^{i+1} - _{W}\textbf{p}_{I}^{i} - _{W}\textbf{v}_{I}^{i}\Delta{t}_i - \frac{1}{2}{_W\textbf{g}}{\Delta{t}_{i}}^2 )\nonumber\\
\boldsymbol{\beta}_{{I}_{i}}^{{i+1}} &=& _{I}\textbf{R}_{W}^{i} (_{W}\textbf{\^{v}}_{I}^{i+1} - _{W}\textbf{v}_{I}^{i} - {_W\textbf{g}}{\Delta{t}_{i}})
\label{eq5}
\end{eqnarray}

Substituting \eq{eq3} into \eq{eq5}, we can estimate $\boldsymbol{\chi}_{S} = [ {\textbf{v}_{{I}}^{{i}}}, {\textbf{v}_{{I}}^{{i+1}}}, _{W}{\textbf{g}}, s_2 ]^{T}$ by solving the linear least square problem in the following form:

\begin{eqnarray}
 \min_{\boldsymbol{\chi}_{S}} \sum_{i \in K}  {\norm{ \hat{\boldsymbol{z}}_{{S}_{i}}^{{i+1}} - \textbf{H}_{{S}_{i}}^{{i+1}}{\boldsymbol{\chi}_{S}}  }^2}
 \label{eq6}
\end{eqnarray}

\noindent where 
$\hat{\boldsymbol{z}}_{{S}_{i}}^{{i+1}} = $
\[
\begin{bmatrix}
\boldsymbol{\hat{\alpha}}_{{I}_{i}}^{{i+1}} - {_{I}\textbf{R}_{W}^{i}}{_{W}\textbf{R}_{C}^{i+1}}{_{C}\textbf{p}_{I}^{i+1}} + {_{I}\textbf{R}_{C}^{i}}{_{C}\textbf{p}_{I}^{i}}\\
\boldsymbol{\hat{\beta}}_{{I}_{i}}^{{i+1}}
\end{bmatrix}\\
\]
and $\textbf{H}_{{S}_{i}}^{{i+1}} = $

\[
\scalebox{.88}{
$\begin{bmatrix}
-\textbf{I}{\Delta{t}_{i}} & \textbf{0} & -\frac{1}{2}{_{I}\textbf{R}_{W}^{i}}{\Delta{t}_{i}}^2 & {_{I}\textbf{R}_{W}^{i}}({_{W}{\textbf{p}_{r1}}_{C}^{i+1}} - {_W{\textbf{p}_{r1}}_{C}^{i}})\\
-\textbf{I} & {_{I}\textbf{R}_{W}^{i}}{_{W}\textbf{R}_{I}^{i+1}} & -{_{I}\textbf{R}_{W}^{i}}{\Delta{t}_{i}} & \textbf{0}
\end{bmatrix}$
}
\]

\subsection{Loop-closing and Relocalization}

In a sliding window and marginalization based optimization method, drift accumulates over time on the pose estimate. A global optimization and relocalization scheme is necessary to eliminate this drift and to achieve global consistency. We adapt DBoW2~\cite{galvez2012bagofwords}, a bag of binary words (BoW) place recognition module, and augment OKVIS for loop detection and relocalization. For each keyframe, only the descriptors of the \textit{keypoints} detected during the local tracking are used to build the BoW database. No new features will be detected in the loop closure step.

A pose-graph is maintained to represent the connection between keyframes. In particular, a node represents a keyframe and an edge between two keyframes exists if the matched keypoints ratio between them is more than $0.75$. In practice, this results into a very sparse graph. With each new keyframe in the pose-graph, the loop-closing module searches for candidates in the bag of words database. A query for detecting loops to the BoW database only returns the candidates outside the current marginalization window and having greater than or equal to score than the neighbor keyframes of that node in the pose-graph. If loop is detected, the candidate with the highest score is retained and feature correspondences between the current keyframe in the local window and the loop candidate keyframes are obtained to establish connection between them. The pose-graph is consequently updated with loop information. A 2D-2D descriptor matching and a 3D-2D matching between the known landmark in the current window keyframe and loop candidate with outlier rejection by PnP RANSAC is performed to obtain the geometric validation.

When a loop is detected, the global relocalization module aligns the current keyframe pose in the local window with the pose of the loop keyframe in the pose-graph by sending back the drift in pose to the windowed sonar-visual-inertial-depth optimization thread. Also, an additional optimization step, similar to \eq{eq7}, is taken only with the matched landmarks with loop candidate for calculating the sonar error term and reprojection error:

\begin{eqnarray}
\textit{J}(\textbf{x}) &=&\sum_{i=1}^{2}\sum_{k=1}^{K}\sum_{j \in \textit{Loop}(i,k)} {\textbf{e}_r^{{i,j,k}^{T}}}{\textbf{P}_r^k}{\textbf{e}_r^{i,j,k}} + \sum_{k=1}^{K-1}{\textbf{e}_t^{k^T}}{\textbf{P}_t^k}{\textbf{e}_t^k}\nonumber
\\
\label{eq8}
\end{eqnarray}

After loop detection, a 6-DoF (position, ${\textbf{x}_{\textbf{p}}}$  and rotation, ${\textbf{x}_{\textbf{q}}}$) pose-graph optimization takes place to optimize over relative constraints between poses to correct drift. The relative transformation between two poses $\textbf{T}_{i}$ and $\textbf{T}_{j}$ for current keyframe in the current window $i$ and keyframe $j$ (either loop candidate keyframe or connected keyframe)  can be calculated from
$\Delta{\textbf{T}_{ij}} = {\textbf{T}_{j}}{\textbf{T}_{i}}^{-1}$. 
The error term, ${\textbf{e}_{{\textbf{x}_{\textbf{p}}}, {\textbf{x}_{\textbf{q}}}}^{i,j}} $ between keyframes $i$ and $j$ is formulated minimally in the tangent space:
\begin{eqnarray}
{\textbf{e}_{{\textbf{x}_{\textbf{p}}}, {\textbf{x}_{\textbf{q}}}}^{i,j}} &=& \Delta{\textbf{T}_{ij}} \hat{{\textbf{T}_{i}}} {\hat{{\textbf{T}_{j}}}}^{-1}
\label{eq9}
\end{eqnarray}

\noindent where ${(\hat{.})}$ denotes the estimated values obtained from local sonar-visual-inertial-depth optimization. The cost function to minimize is given by
\begin{eqnarray}
\textit{J}({\textbf{x}_{\textbf{p}}}, {\textbf{x}_{\textbf{q}}}) &=& \sum_{i,j} {\textbf{e}_{{\textbf{x}_{\textbf{p}}}, {\textbf{x}_{\textbf{q}}}}^{i,j}}^T {\textbf{P}_{{\textbf{x}_{\textbf{p}}}, {\textbf{x}_{\textbf{q}}}}^{i,j}} {\textbf{e}_{{\textbf{x}_{\textbf{p}}}, {\textbf{x}_{\textbf{q}}}}^{i,j}}\nonumber\\
&+& \sum_{(i,j) \in \textit{Loop}} \rho({\textbf{e}_{{\textbf{x}_{\textbf{p}}}, {\textbf{x}_{\textbf{q}}}}^{i,j}}^T {\textbf{P}_{{\textbf{x}_{\textbf{p}}}, {\textbf{x}_{\textbf{q}}}}^{i,j}} {\textbf{e}_{{\textbf{x}_{\textbf{p}}}, {\textbf{x}_{\textbf{q}}}}^{i,j}})
 \label{eq10}
\end{eqnarray}

\noindent where ${\textbf{P}_{{\textbf{x}_{\textbf{p}}}, {\textbf{x}_{\textbf{q}}}}^{i,j}}$ is the information matrix set to identity, as in~\cite{strasdat2012local}, and $\rho$ is the Huber loss function to potentially down-weigh any  incorrect loops.

\section{EXPERIMENTAL RESULTS}
\label{sec:results}
The proposed state estimation system, SVIn2, is quantitatively validated first on a standard dataset, to ensure that loop closure and the initialization work also above water. Moreover, it is compared to other state-of-the-art methods, i.e., VINS-Mono~\cite{qin2018vins}, the basic OKVIS~\cite{okvis}, and the MSCKF~\cite{mourikis2007multi} implementation from the GRASP lab~\cite{msckf-danilidis}.
Second, we qualitatively test the proposed approach on several different datasets collected utilizing a custom made sensor suite~\cite{RahmanOceans2018} and an Aqua2 AUV~\cite{Rekleitis2005d}.

\subsection{Validation on Standard dataset}

Here, we present results on the EuRoC dataset~\cite{Burri25012016}, one of the benchmark datasets used by many visual\hyp inertial state estimation systems, including OKVIS (Stereo), VINS-Mono, and MSCKF.  To compare the performance, we disable depth and sonar integration in our method and only assess the loop-closure scheme.

\begin{figure}
 \centering
 \includegraphics[trim={5.5cm 0 5.5cm 0},clip,width=.85\columnwidth]{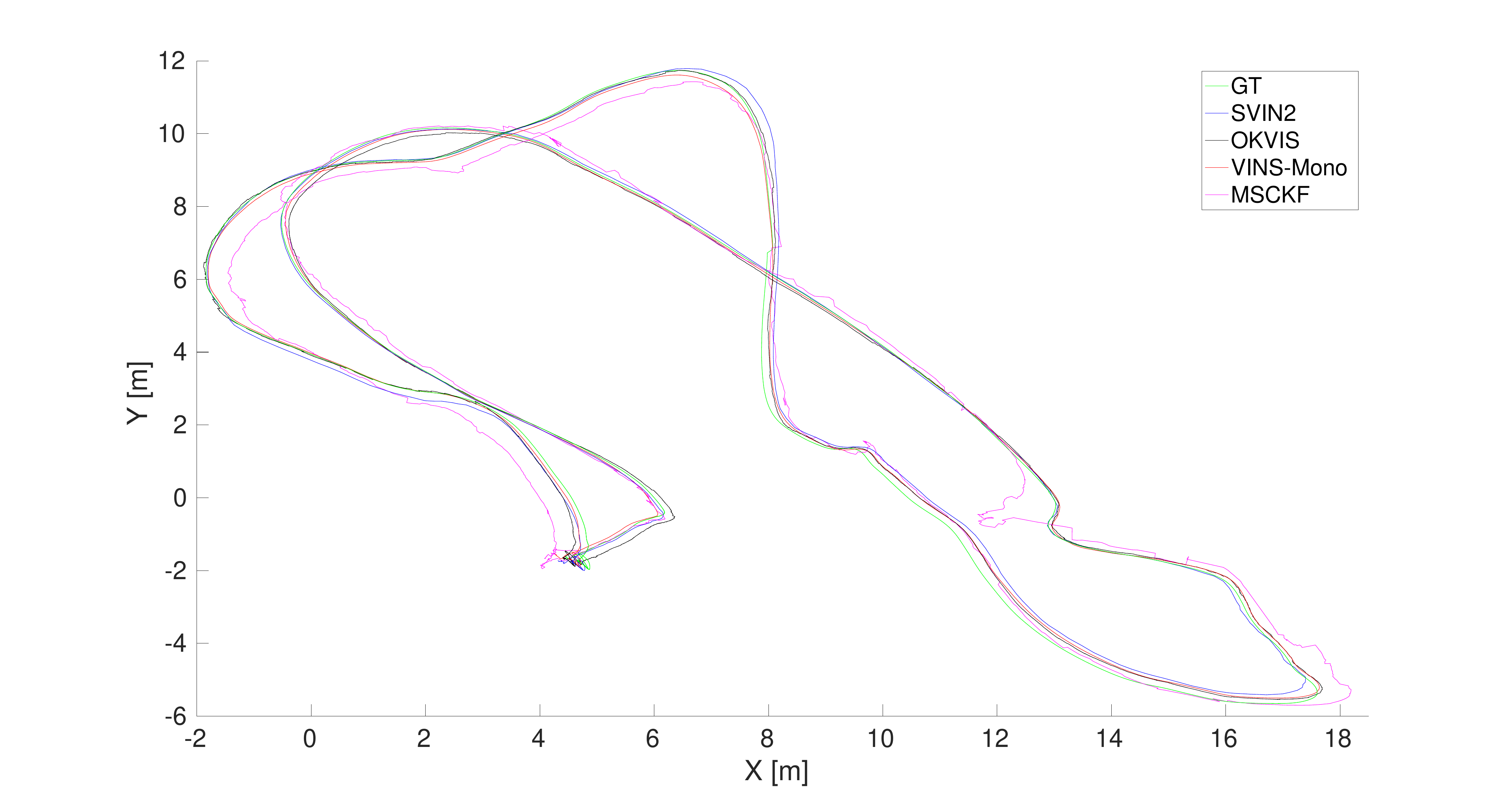}
 \vspace{-0.1in}\caption{Trajectories on the MH 04 sequence of the EuRoC dataset. \vspace{-0.4in} \label{fig:euroc-results}}
\end{figure}

Following the current benchmarking practices, an alignment is performed between ground truth and estimated trajectory, by minimizing the least mean square errors between estimate/ground\hyp truth locations, which are temporally close, varying rotation and translation, according to the method from \cite{Umeyama:1991:LET:105514.105525}.
The resulting metric is the Root Mean Square Error (RMSE) for the translation, shown in \tab{tab:euroc-rmse} for several Machine Hall sequences in the EuRoC dataset. For each package, every sequence has been run 5 times and the best run (according to RMSE) has been shown. Our method shows reduced RMSE in every sequence from OKVIS, validating the improvement of pose-estimation after loop-closing. SVIn2 has also less RMSE than MSCKF and slightly higher in some sequences, but comparable, to results from VINS-Mono.
\fig{fig:euroc-results} shows the trajectories for each method together with the ground truth for the \textit{Machine Hall 04 Difficult} sequence.

\begin{table}[t]
\begin{center}
\caption{The best absolute trajectory error (RMSE) in meters for each Machine Hall EuRoC sequence.\label{tab:euroc-rmse}}
\begin{tabular}[t]{@{} l|c|c|c|c|}
 & \rotatebox{90}{SVIn2} & \rotatebox{90}{OKVIS(stereo)} & \rotatebox{90}{VINS-Mono} & \rotatebox{90}{MSCKF}\\
\midrule
   MH 01 & 0.13 & 0.15 & 0.07 & 0.21 \\
   MH 02 & 0.08 & 0.14 & 0.08 & 0.24 \\
   MH 03 & 0.07 & 0.12 & 0.05 & 0.24 \\
   MH 04 & 0.13 & 0.18 & 0.15 & 0.46 \\
   MH 05 & 0.15 & 0.24 & 0.11 & 0.54 \\
\bottomrule
\end{tabular}
\end{center}\vspace{-0.2in}
\end{table}

\subsection{Underwater datasets}
Our proposed state estimation system -- SVIn2 -- is targeted for the underwater environment, where sonar and depth can be fused together with the visual\hyp inertial data. The stereo cameras are configured to capture frames at $15$ \si{fps}, IMU at $100$ \si{Hz}, Sonar at $100$ \si{Hz}, and Depth sensor at $1$ \si{Hz}.
Here, we show results from four different datasets in three different underwater environments. First, a sunken bus in Fantasy Lake (NC), where data was collected by a diver with a custom\hyp made underwater sensor suite \cite{RahmanOceans2018}. The diver started from outside the bus, performed a loop around and entered in it from the back door, exited across and finished at the front-top of the bus.
The images are affected by haze and low visibility.
Second and third, data from an underwater cavern in Ginnie Springs (FL) is collected again by a diver with the same sensor suite as for the sunken bus. The diver performed several loops, around one spot in the second dataset -- Cavern1 -- and two spots in the third dataset -- Cavern2 -- inside the cavern. The environment is affected by complete absence of natural light.
Fourth, an AUV -- Aqua2 robot -- collected data over a fake underwater cemetery in Lake Jocassee (SC) and performed several loops around the tombstones in a square pattern. The visibility, as well as brightness and contrast, was very low. 
In the underwater datasets, it is a challenge to get any ground truth, because it is a GPS-denied unstructured environment. As such, the evaluation is qualitative, with a rough estimate on the size of the environment measured beforehand by the divers collecting the data.

\begin{figure}
\centering
{\includegraphics[width=0.8\columnwidth]{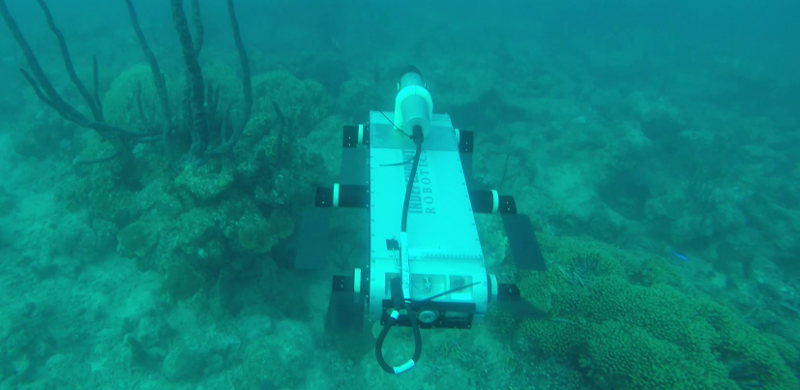}}
\vspace{-0.1in}\caption{The Aqua2 AUV~\cite{Rekleitis2005d} equipped with the scanning sonar collecting data over the coral reef.\vspace{-0.2in}}
\label{fig:aqua}
\end{figure}

\begin{figure*}[h]
\begin{center}
\leavevmode
\begin{tabular}{c}
 \subfigure[]{\includegraphics[trim=0.0in 0.0in 0.0in 0.4in,clip,height=1.45in]{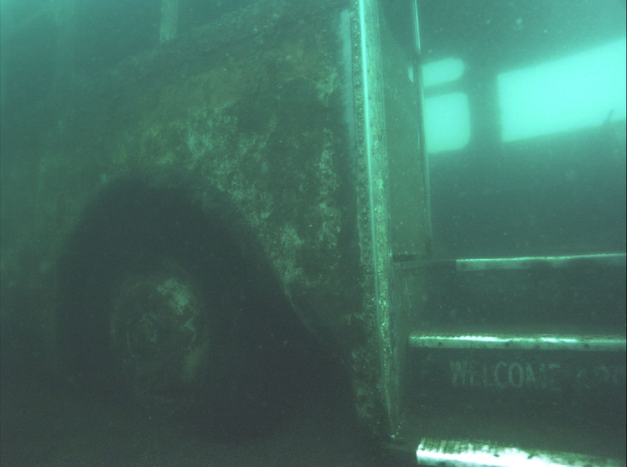}\label{fig:EnvA}}
 \subfigure[]{\includegraphics[trim={3.0cm 9cm 9.0cm 2cm},clip,height=1.45in]{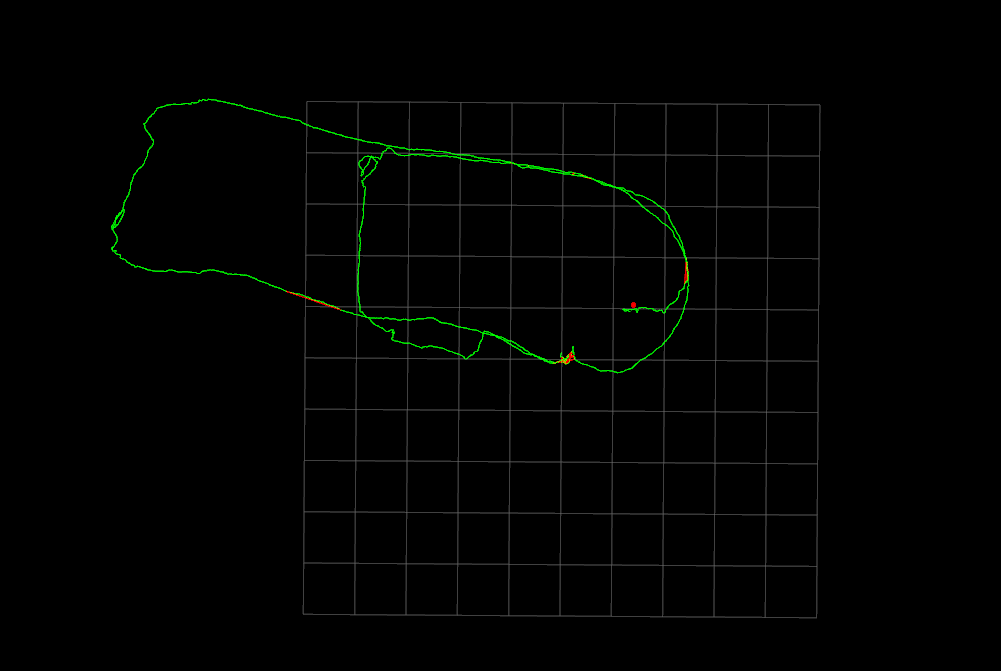}\label{fig:EnvA-rviz}}
 \subfigure[]{\includegraphics[trim={0.0cm 0 0.0cm 0},clip,height=1.45in]{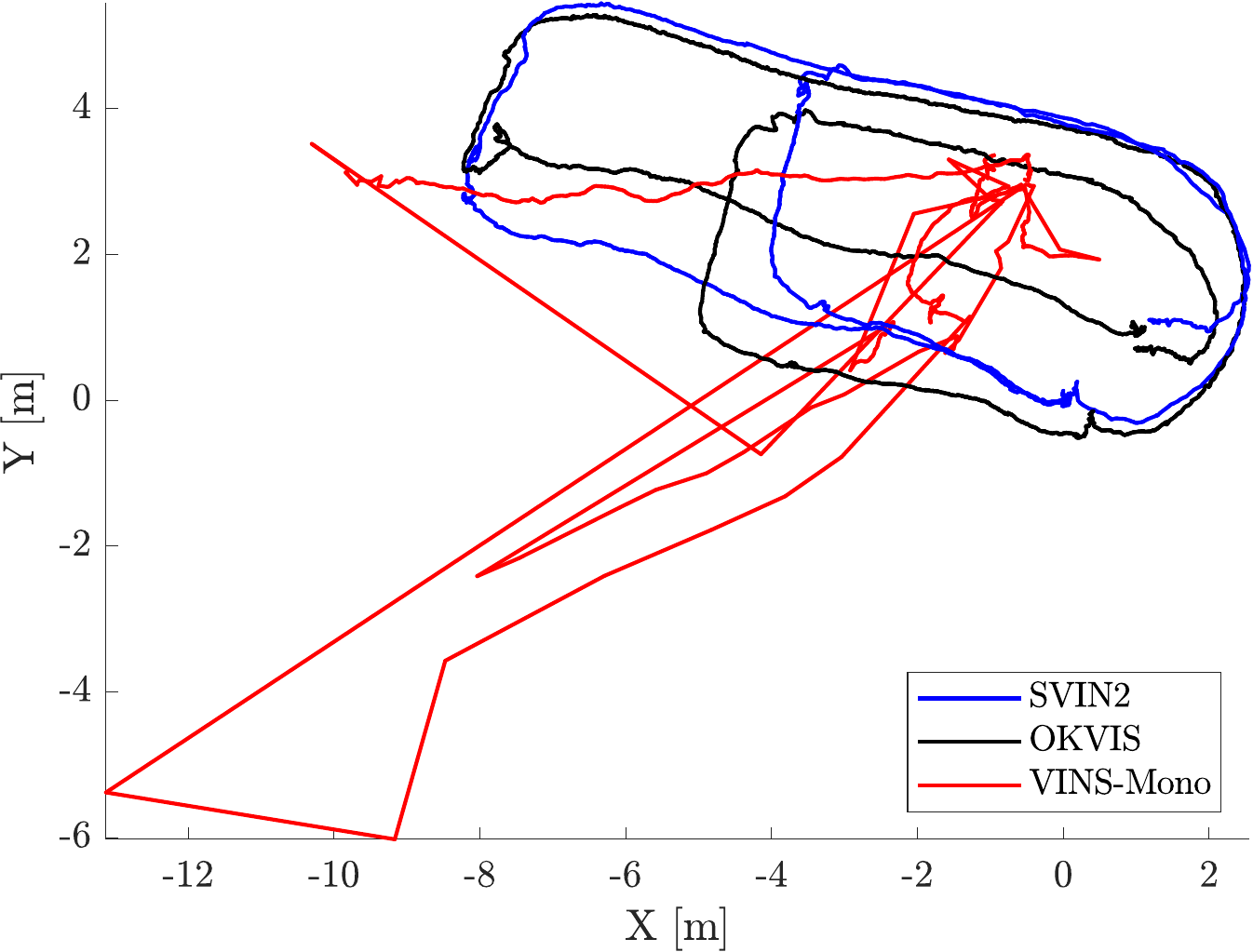}\label{fig:EnvA-matlab}}
\end{tabular}
\end{center}
\vspace{-0.1in}\caption{\subref{fig:EnvA} Submerged bus, Fantasy Lake, NC, USA with a $53$ \si{m} trajectory; trajectories from SVIn2 with all sensors enabled shown in rviz \subref{fig:EnvA-rviz} and aligned trajectories from SVIn2 with Sonar and depth disabled, OKVIS, and VINS-Mono \subref{fig:EnvA-matlab} are displayed.}
   \label{fig:EnvBus}
 \end{figure*}

\begin{figure*}[h]
\begin{center}
\leavevmode
\begin{tabular}{c}
\subfigure[]{\includegraphics[trim=0.0in 0.0in 0.0in 0.4in,clip,height=1.45in]{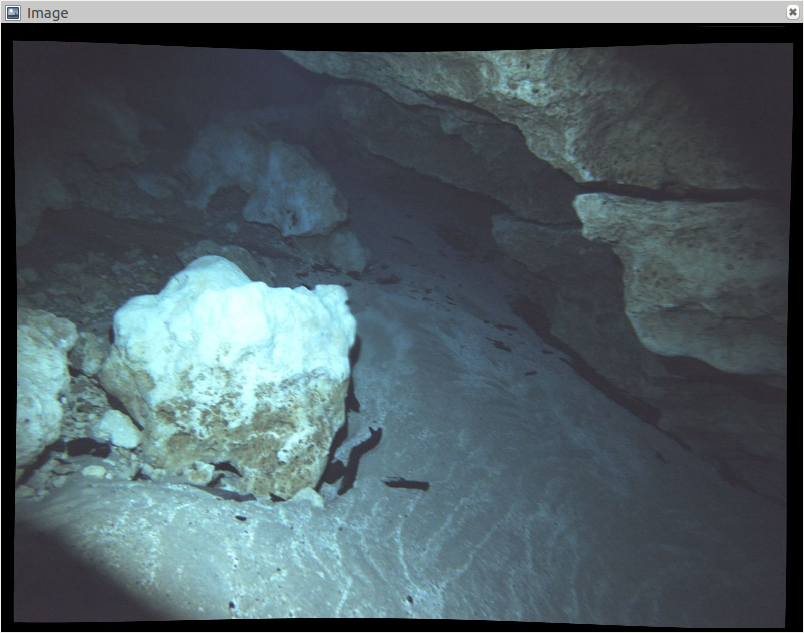}\label{fig:EnvB}}
\subfigure[]{\includegraphics[trim={8cm 5cm 0cm 0},clip,height=1.45in]{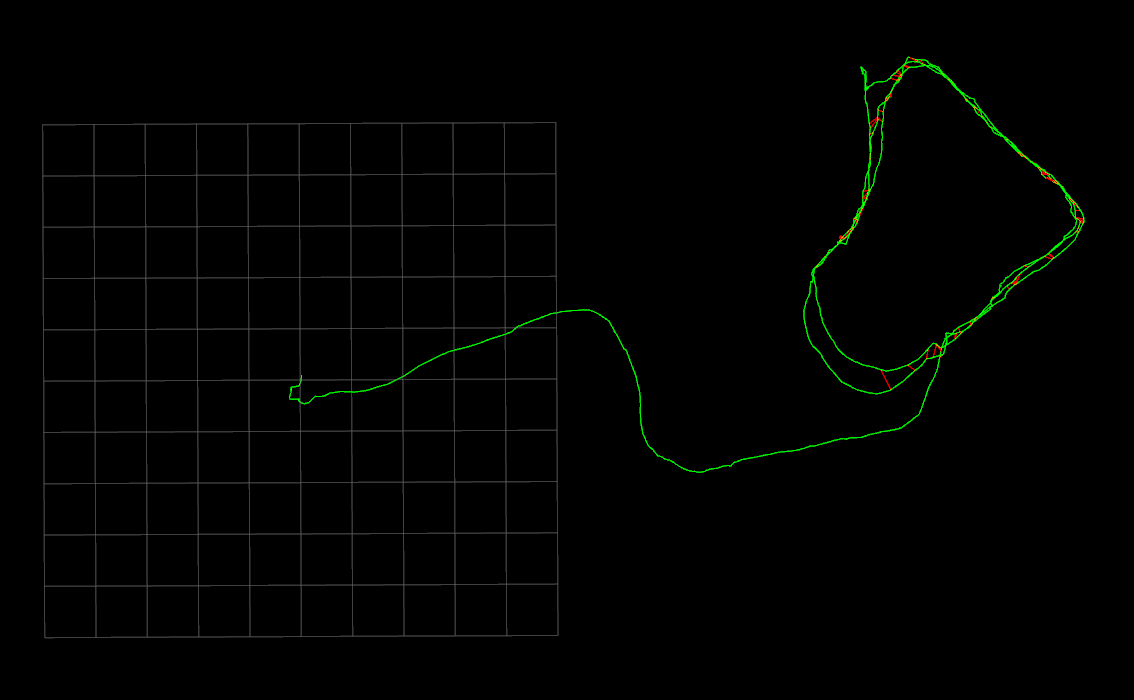}\label{fig:EnvB-rviz}}
\subfigure[]{\includegraphics[trim={0cm 0 0cm 0},clip,height=1.35in]{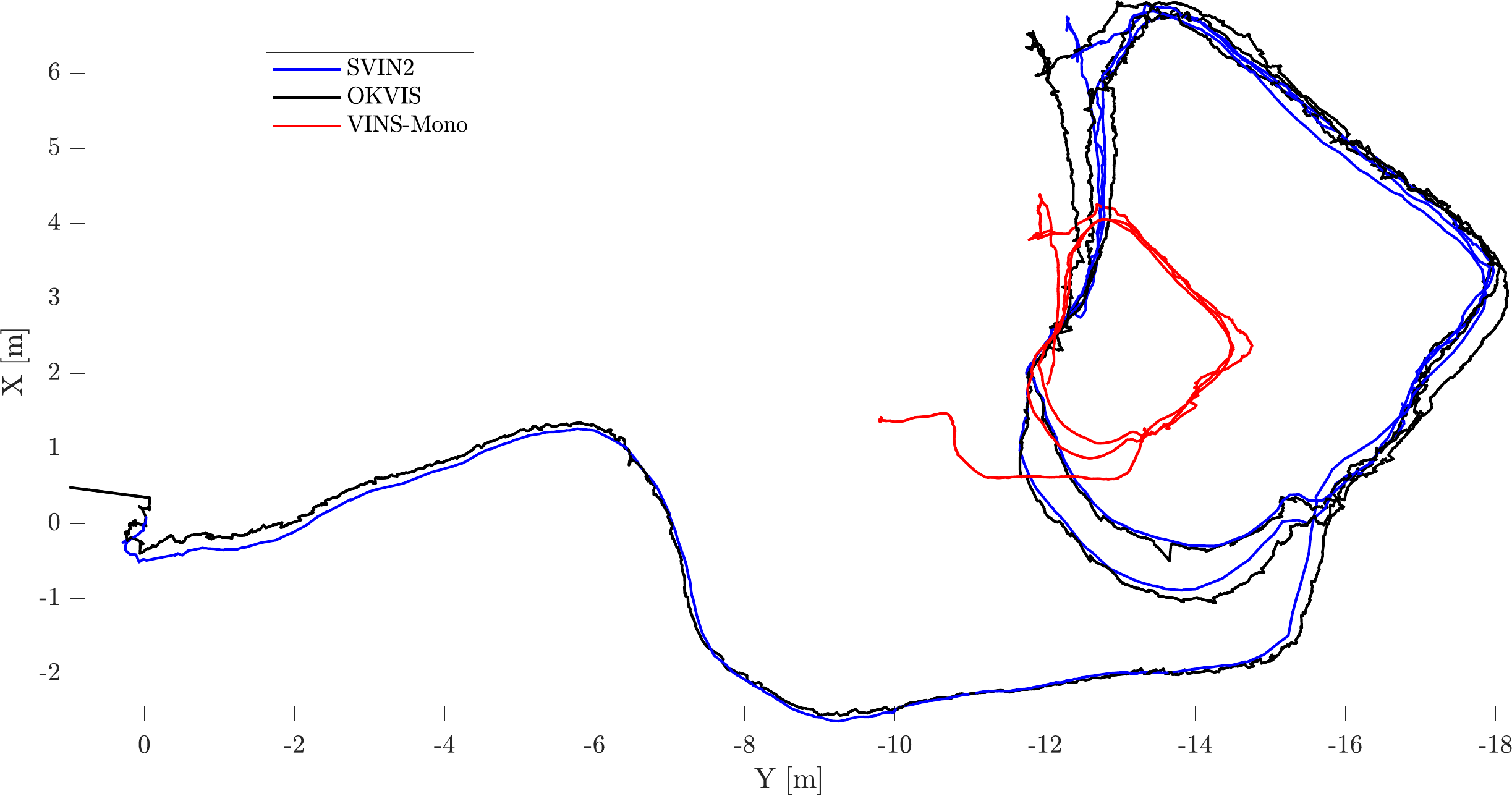}\label{fig:EnvB-matlab}}\\
\end{tabular}
\end{center}
\vspace{-0.1in}\caption{\subref{fig:EnvB} Cave environment, Ballroom, Ginnie Springs, FL, USA, with a unique loop covering a $87$ \si{m} trajectory;  trajectories from SVIn2 with all sensors enabled shown in rviz \subref{fig:EnvB-rviz} and aligned trajectories from SVIn2 with Sonar and depth disabled, OKVIS, and VINS-Mono \subref{fig:EnvB-matlab} are displayed.}
   \label{fig:EnvCave1}
 \end{figure*}

\begin{figure*}[h]
\begin{center}
\leavevmode
\begin{tabular}{c}
\subfigure[]{\includegraphics[trim=0.0in 0.0in 0.0in 0.4in,clip,height=1.4in]{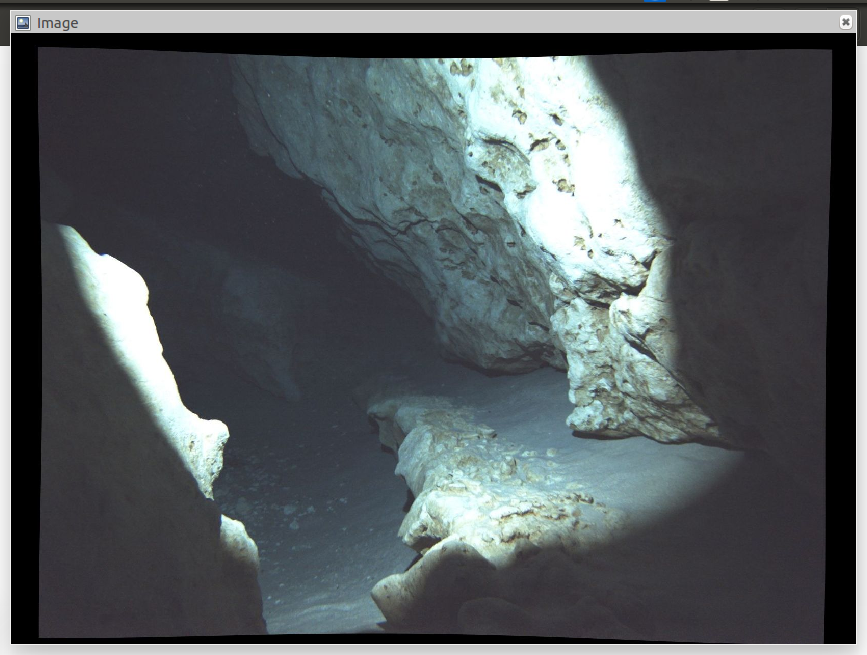}\label{fig:EnvC}}
\subfigure[]{\includegraphics[trim={8cm 5cm 0cm 0},clip,height=1.4in]{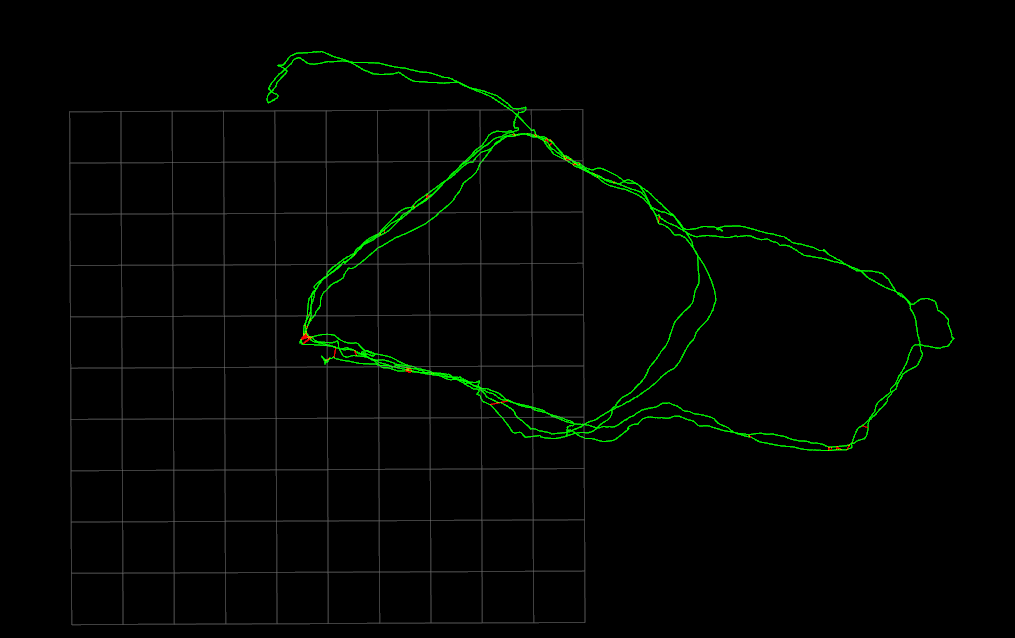}\label{fig:EnvC-rviz}}
\subfigure[]{\includegraphics[trim={0cm 0cm 0cm 0},clip,height=1.4in]{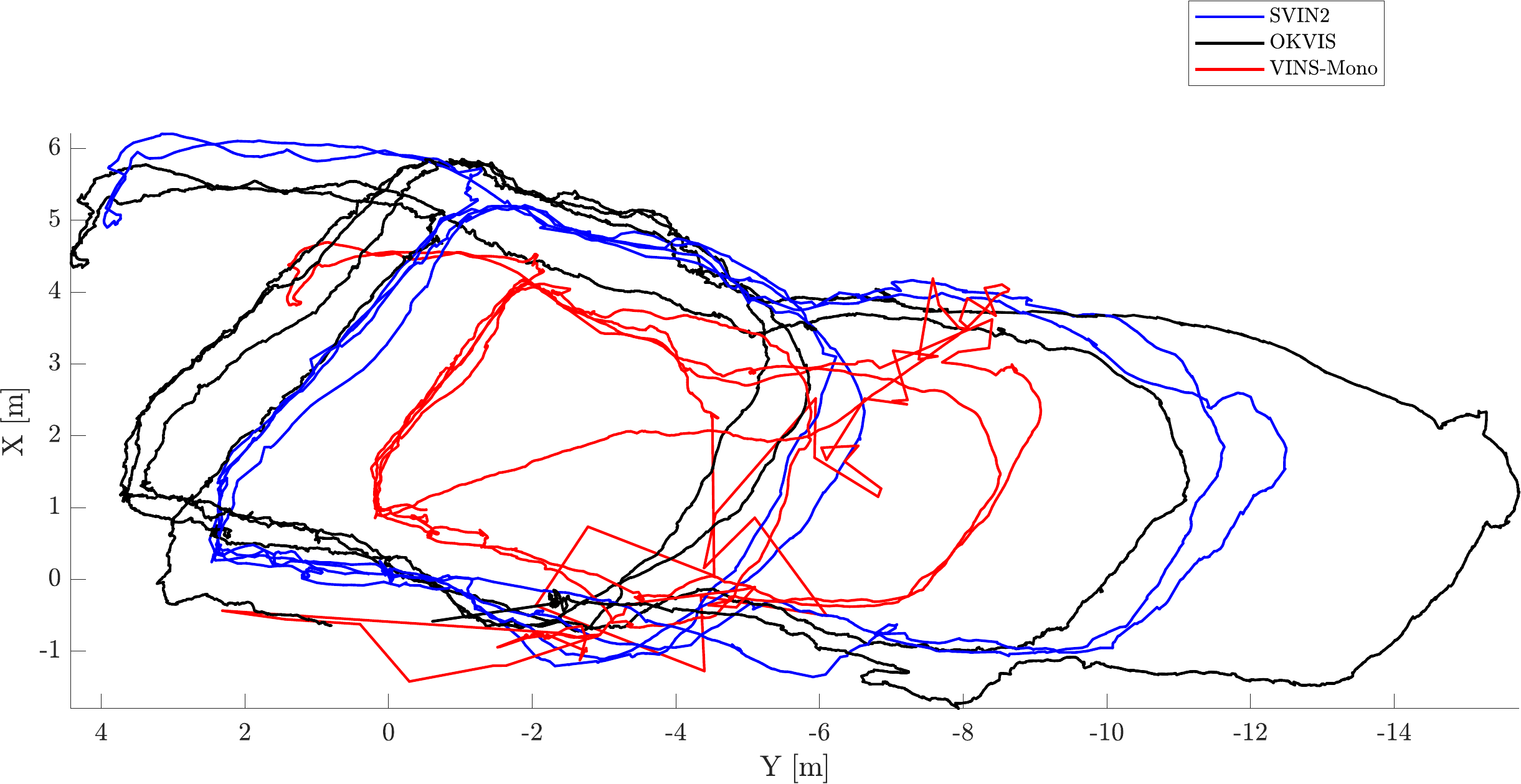}\label{fig:EnvC-matlab}}
\end{tabular}
\end{center}
\vspace{-0.1in}\caption{\subref{fig:EnvC} Cave environment, Ballroom, Ginnie Springs, FL, USA, with two loops in different areas covering a $155$ \si{m} trajectory;  trajectories from SVIn2 with all sensors enabled shown in rviz \subref{fig:EnvC-rviz} and aligned trajectories from SVIn2 with Sonar and depth disabled, OKVIS, and VINS-Mono \subref{fig:EnvC-matlab} are displayed.}
   \label{fig:EnvCave2}
 \end{figure*}

\begin{figure*}[h]
\begin{center}
\leavevmode
\begin{tabular}{c}
\subfigure[]{\includegraphics[trim=0.0in 0.0in 0.0in 0.4in,clip,height=1.45in]{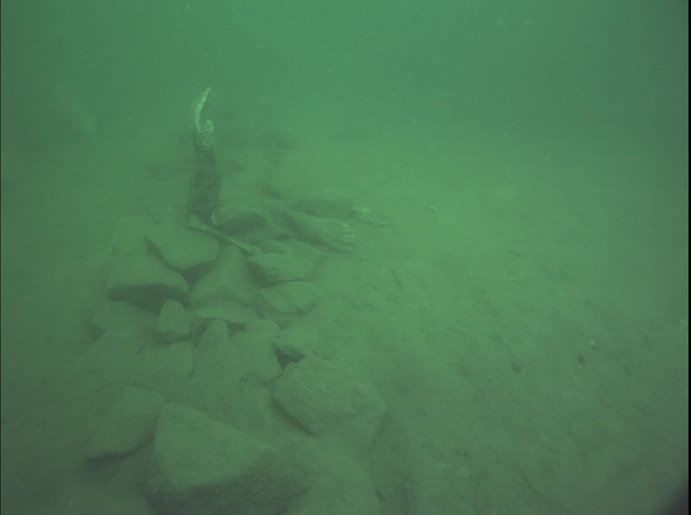}\label{fig:EnvD}}
\subfigure[]{\includegraphics[trim={0.0cm 0 0.0cm 0},clip,height=1.45in]{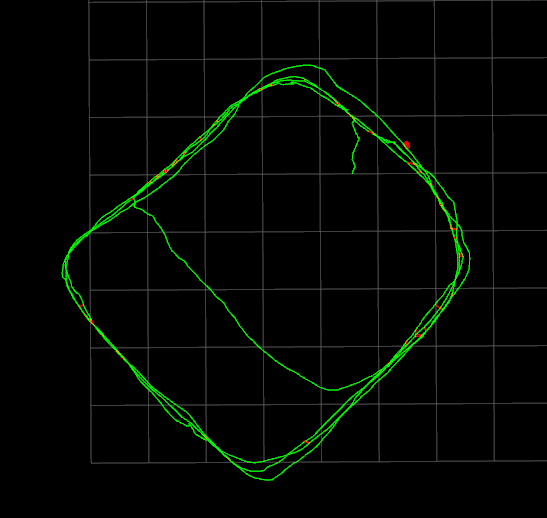}\label{fig:EnvD-rviz}}
\subfigure[]{\includegraphics[trim={0.0cm 0 0.0cm 0},clip,height=1.45in]{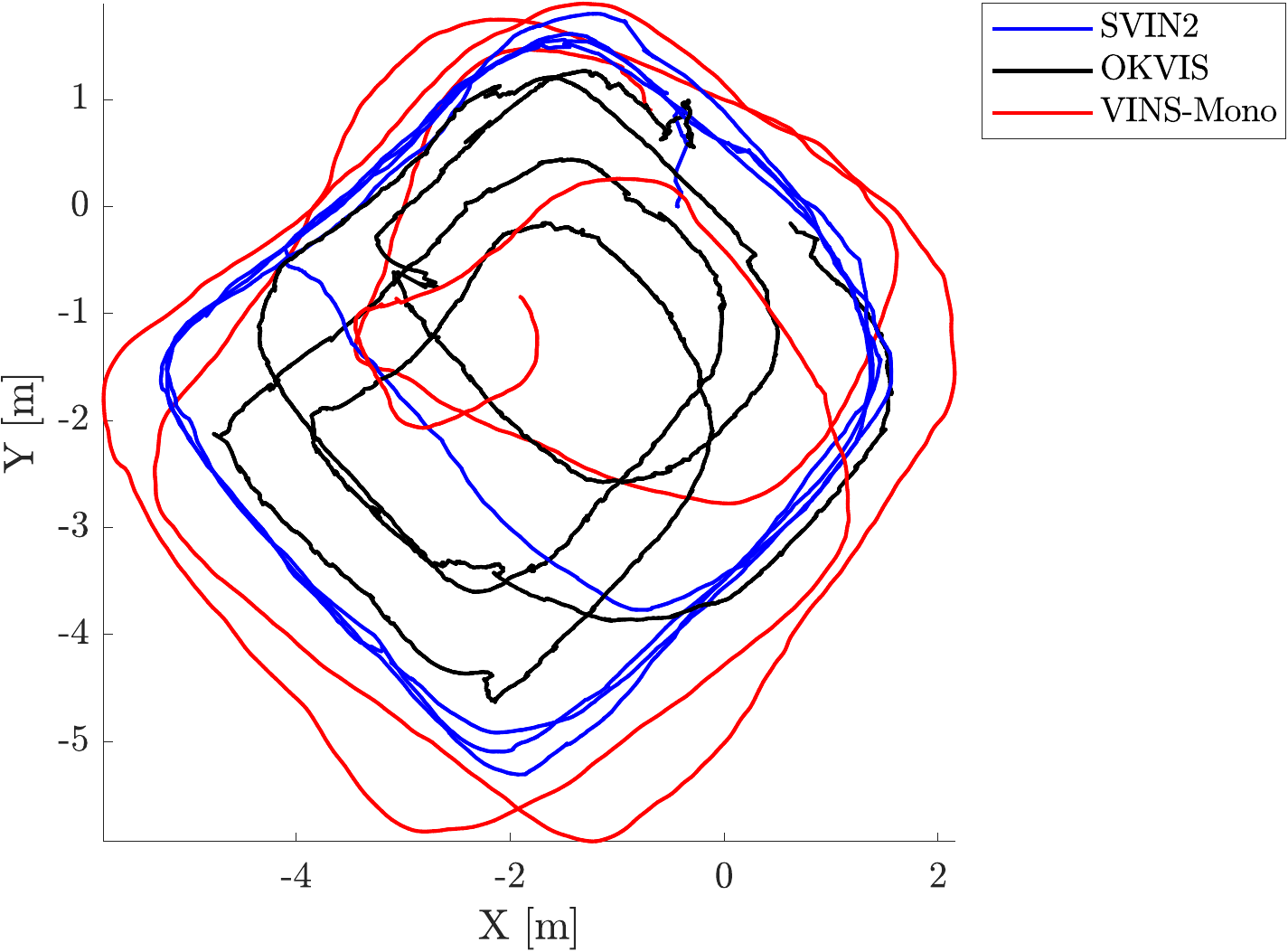}\label{fig:EnvD-matlab}}
\end{tabular}
\end{center}
\vspace{-0.1in}\caption{\subref{fig:EnvD} Aqua2 in a fake cemetery, Lake Jocassee, SC, USA with a $80$ \si{m} trajectory;  trajectories from SVIn2 with visual, inertial, and depth sensor (no sonar data has been used) shown in rviz \subref{fig:EnvD-rviz} and aligned trajectories from SVIn2 with Sonar and depth disabled, OKVIS, and VINS-Mono \subref{fig:EnvD-matlab} are displayed.}
   \label{fig:Env}
 \end{figure*}

Figs. \ref{fig:EnvBus}-\ref{fig:Env} show the trajectories from SVIn2, OKVIS, and VINS-Mono in the datasets just described. MSCKF was able to keep track only for some small segments in all the datasets, hence excluded from the plots. For a fair comparison, when the trajectories were compared against each other, sonar and depth were disabled in SVIn2. All trajectories are plotted keeping the original scale produced by each package. 

\fig{fig:EnvBus} shows the results for the submerged bus dataset.  VINS-Mono lost track when the exposure increased for quite some time. It tried to re-initialize, but it was not able to track successfully. Even using \textit{histogram equalization} or a \textit{contrast adjusted histogram equalization} filter, VINS-Mono was not able to track. Even if the scale drifted, OKVIS was able to track using a \textit{contrast adjusted histogram equalization} filter in the image pre-processing step. Without the filter, it lost track at the high exposure location. The proposed method was able to track, detect, and correct the loop, successfully.

In Cavern1 -- see \fig{fig:EnvCave1} -- VINS-Mono tracked successfully the whole time. However, as can be noticed in \fig{fig:EnvB-matlab}, the scale was incorrect based on empirical observations during data collection. OKVIS instead produced a good trajectory, and SVIn2 was also able to detect and close the loops.

In Cavern2 (\fig{fig:EnvCave2}), VINS-Mono lost track at the beginning, reinitialized, was able to track for some time, and detected a loop, before losing track again. VINS-Mono had similar behavior even if the images were pre-processed with different filters. OKVIS tracked well, but as drifts accumulated over time, it was not able to join the current pose with a previous pose where a loop was expected. SVIn2 was able to track and reduce the drift in the trajectory with successful loop closure.

In the cemetery dataset -- \fig{fig:Env} -- both VINS-Mono and OKVIS were able to track, but VINS-Mono was not able to reduce the drift in trajectory, while SVIn2 was able to fuse and correct the loops.

\section{CONCLUSIONS}
In this paper, we presented SVIn2, a state estimation system with robust initialization, sensor fusion of depth, sonar, visual, and inertial data, and loop closure capabilities. While the proposed system can also work out of the water, by disabling the sensors that are not applicable, our system is specifically targeted for underwater environments. Experimental results in a standard benchmark dataset and different underwater datasets demonstrate excellent performance.

Utilizing the insights gained from implementing the proposed approach, an online adaptation of the discussed framework for the limited computational resources of the Aqua2 AUV~\cite{Rekleitis2005d} is currently under consideration; see \fig{fig:aqua}. It is worth noting that maintaining the proper attitude of the traversed trajectory and providing an estimate of the distance traveled will greatly enhance the autonomous capabilities of the vehicle~\cite{RekleitisIROS2009b}. Furthermore, accurately modeling the surrounding structures would enable Aqua2, as well as other vision based underwater vehicles to operate near, and through, a variety of underwater structures, such as caves, shipwrecks, and canyons.


\bibliographystyle{IEEEtran}
\bibliography{IEEEabrv,refs}

\begin{thebibliography}{10}
\providecommand{\url}[1]{#1}
\csname url@rmstyle\endcsname
\providecommand{\newblock}{\relax}
\providecommand{\bibinfo}[2]{#2}
\providecommand\BIBentrySTDinterwordspacing{\spaceskip=0pt\relax}
\providecommand\BIBentryALTinterwordstretchfactor{4}
\providecommand\BIBentryALTinterwordspacing{\spaceskip=\fontdimen2\font plus
\BIBentryALTinterwordstretchfactor\fontdimen3\font minus
  \fontdimen4\font\relax}
\providecommand\BIBforeignlanguage[2]{{%
\expandafter\ifx\csname l@#1\endcsname\relax
\typeout{** WARNING: IEEEtran.bst: No hyphenation pattern has been}%
\typeout{** loaded for the language `#1'. Using the pattern for}%
\typeout{** the default language instead.}%
\else
\language=\csname l@#1\endcsname
\fi
#2}}

\bibitem{ballard2014smithsonian}
R.~Ballard, ``Why we must explore the sea,'' \emph{Smithsonian Magazine}, 2014.

\bibitem{henderson2013mapping}
J.~Henderson, O.~Pizarro, M.~Johnson-Roberson, and I.~Mahon, ``Mapping
  submerged archaeological sites using stereo-vision photogrammetry,''
  \emph{International Journal of Nautical Archaeology}, vol.~42, no.~2, pp.
  243--256, 2013.

\bibitem{leonard2012directed}
J.~J. Leonard and H.~F. Durrant-Whyte, \emph{Directed sonar sensing for mobile
  robot navigation}.\hskip 1em plus 0.5em minus 0.4em\relax Springer Science \&
  Business Media, 2012, vol. 175.

\bibitem{lee2005underwater}
C.-M. Lee \emph{et~al.}, ``Underwater navigation system based on inertial
  sensor and doppler velocity log using indirect feedback {Kalman} filter,''
  \emph{International Journal of Offshore and Polar Engineering}, vol.~15,
  no.~02, 2005.

\bibitem{snyder2010doppler}
J.~Snyder, ``{Doppler Velocity Log (DVL)} navigation for observation-class
  {ROVs},'' in \emph{MTS/IEEE OCEANS, SEATTLE}, 2010, pp. 1--9.

\bibitem{johannsson2010imaging}
H.~Johannsson, M.~Kaess, B.~Englot, F.~Hover, and J.~Leonard, ``Imaging
  sonar-aided navigation for autonomous underwater harbor surveillance,'' in
  \emph{IEEE/RSJ International Conference on Intelligent Robots and Systems
  (IROS)}.\hskip 1em plus 0.5em minus 0.4em\relax IEEE, 2010, pp. 4396--4403.

\bibitem{rigby2006towards}
P.~Rigby, O.~Pizarro, and S.~B. Williams, ``Towards geo-referenced {AUV}
  navigation through fusion of {USBL} and {DVL} measurements,'' in
  \emph{OCEANS}, 2006, pp. 1--6.

\bibitem{mur2017visual}
R.~Mur-Artal and J.~D. Tard{\'o}s, ``Visual-inertial monocular {SLAM} with map
  reuse,'' \emph{{IEEE} Robot. Autom. Lett.}, vol.~2, no.~2, pp. 796--803,
  2017.

\bibitem{okvis}
S.~Leutenegger, S.~Lynen, M.~Bosse, R.~Siegwart, and P.~Furgale,
  ``Keyframe-based visual-inertial odometry using nonlinear optimization,''
  \emph{Int. J. Robot. Res.}, vol.~34, no.~3, pp. 314--334, 2015.

\bibitem{qin2018vins}
T.~Qin, P.~Li, and S.~Shen, ``{VINS-Mono}: A robust and versatile monocular
  visual-inertial state estimator,'' \emph{{IEEE} Trans. Robot.}, vol.~34,
  no.~4, pp. 1004--1020, 2018.

\bibitem{mourikis2007multi}
A.~I. Mourikis and S.~I. Roumeliotis, ``A multi-state constraint {Kalman}
  filter for vision-aided inertial navigation,'' in \emph{Proc. ICRA}.\hskip
  1em plus 0.5em minus 0.4em\relax IEEE, 2007, pp. 3565--3572.

\bibitem{msckf-kumar-ral}
K.~Sun, K.~Mohta, B.~Pfrommer, M.~Watterson, S.~Liu, Y.~Mulgaonkar, C.~J.
  Taylor, and V.~Kumar, ``Robust stereo visual inertial odometry for fast
  autonomous flight,'' \emph{IEEE Robot. Autom. Lett.}, vol.~3, no.~2, pp.
  965--972, 2018.

\bibitem{RekleitisISERVO2016}
A.~{Quattrini Li}, A.~Coskun, S.~M. Doherty, S.~Ghasemlou, A.~S. Jagtap,
  M.~Modasshir, S.~Rahman, A.~Singh, M.~Xanthidis, J.~M. O'Kane, and
  I.~Rekleitis, ``Experimental comparison of open source vision based state
  estimation algorithms,'' in \emph{Proc. ISER}, 2016.

\bibitem{rahman2018icra}
S.~Rahman, A.~{Quattrini Li}, and I.~Rekleitis, ``{Sonar Visual Inertial SLAM
  of Underwater Structures},'' in \emph{Proc. ICRA}, 2018.

\bibitem{Burri25012016}
M.~Burri, J.~Nikolic, P.~Gohl, T.~Schneider, J.~Rehder, S.~Omari, M.~W.
  Achtelik, and R.~Siegwart, ``The {EuRoC} micro aerial vehicle datasets,''
  \emph{Int. J. Robot. Res.}, vol.~35, no.~10, pp. 1157--1163, 2016.

\bibitem{RahmanOceans2018}
S.~Rahman, A.~{Quattrini Li}, and I.~Rekleitis, ``A modular sensor suite for
  underwater reconstruction,'' in \emph{MTS/IEEE Oceans Charleston}, 2018, pp.
  1--6.

\bibitem{Rekleitis2005d}
G.~Dudek, M.~Jenkin, C.~Prahacs, A.~Hogue, J.~Sattar, P.~Giguere, A.~German,
  H.~Liu, S.~Saunderson, A.~Ripsman, S.~Simhon, L.~A. Torres-Mendez, E.~Milios,
  P.~Zhang, and I.~Rekleitis, ``A visually guided swimming robot,'' in
  \emph{Proc. IROS}, 2005, pp. 1749--1754.

\bibitem{joshi2019iros}
B.~Joshi, S.~Rahman, M.~Kalaitzakis, B.~Cain, J.~Johnson, M.~Xanthidis,
  N.~Karapetyan, A.~Hernandez, A.~{Quattrini Li}, N.~Vitzilaios, and
  I.~Rekleitis, ``{Experimental Comparison of Open Source Visual-Inertial-Based
  State Estimation Algorithms in the Underwater Domain},'' in \emph{Proc.
  IROS}, 2019, (accepted).

\bibitem{folkesson2007feature}
J.~Folkesson, J.~Leonard, J.~Leederkerken, and R.~Williams, ``Feature tracking
  for underwater navigation using sonar,'' in \emph{Proc. IROS}.\hskip 1em plus
  0.5em minus 0.4em\relax IEEE, 2007, pp. 3678--3684.

\bibitem{fallon2013relocating}
M.~F. Fallon, J.~Folkesson, H.~McClelland, and J.~J. Leonard, ``Relocating
  underwater features autonomously using sonar-based {SLAM},'' \emph{IEEE J.
  Oceanic Eng.}, vol.~38, no.~3, pp. 500--513, 2013.

\bibitem{richmond2018sunfish}
K.~Richmond, C.~Flesher, L.~Lindzey, N.~Tanner, and W.~C. Stone,
  ``{SUNFISH}{\textregistered}: A human-portable exploration {AUV} for complex
  {3D} environments,'' in \emph{MTS/IEEE OCEANS Charleston}, 2018, pp. 1--9.

\bibitem{salvi2008visual}
J.~Salvi, Y.~Petillo, S.~Thomas, and J.~Aulinas, ``Visual {SLAM} for underwater
  vehicles using video velocity log and natural landmarks,'' in \emph{MTS/IEEE
  OCEANS}, 2008, pp. 1--6.

\bibitem{beall2011bundle}
C.~Beall, F.~Dellaert, I.~Mahon, and S.~B. Williams, ``Bundle adjustment in
  large-scale 3d reconstructions based on underwater robotic surveys,'' in
  \emph{MTS/IEEE OCEANS, Spain}, 2011, pp. 1--6.

\bibitem{shkurti2011state}
F.~Shkurti, I.~Rekleitis, M.~Scaccia, and G.~Dudek, ``State estimation of an
  underwater robot using visual and inertial information,'' in \emph{Proc.
  IROS}, 2011, pp. 5054--5060.

\bibitem{corke2007experiments}
P.~Corke, C.~Detweiler, M.~Dunbabin, M.~Hamilton, D.~Rus, and I.~Vasilescu,
  ``Experiments with underwater robot localization and tracking,'' in
  \emph{Proc. ICRA}.\hskip 1em plus 0.5em minus 0.4em\relax IEEE, 2007, pp.
  4556--4561.

\bibitem{Civera2010}
J.~Civera, O.~G. Grasa, A.~J. Davison, and J.~M.~M. Montiel, ``{1Point RANSAC
  for Extended Kalman Filtering: Application to Real-time Structure from Motion
  and Visual Odometry},'' \emph{J. Field Robot.}, vol.~27, no.~5, pp. 609--631,
  2010.

\bibitem{4538852}
G.~Klein and D.~Murray, ``Parallel tracking and mapping for small {AR}
  workspaces,'' in \emph{IEEE and ACM Int. Symp. on Mixed and Augmented
  Reality}, 2007, pp. 225--234.

\bibitem{ORBSLAM2015}
R.~Mur-Artal, J.~M.~M. Montiel, and J.~D. Tard\'os, ``{ORB-SLAM: A Versatile
  and Accurate Monocular SLAM System},'' \emph{IEEE Trans. Robot.}, vol.~31,
  no.~5, pp. 1147--1163, 2015.

\bibitem{raey}
J.~Engel, T.~Schöps, and D.~Cremers, ``\BIBforeignlanguage{English}{{LSD-SLAM:
  Large-Scale Direct Monocular SLAM}},'' in
  \emph{\BIBforeignlanguage{English}{Proc. ECCV}}, D.~Fleet, T.~Pajdla,
  B.~Schiele, and T.~Tuytelaars, Eds.\hskip 1em plus 0.5em minus 0.4em\relax
  Springer, 2014, vol. 8690, pp. 834--849.

\bibitem{engel2018direct}
J.~Engel, V.~Koltun, and D.~Cremers, ``Direct sparse odometry,'' \emph{{IEEE}
  Trans. Pattern Anal. Mach. Intell.}, vol.~40, no.~3, pp. 611--625, 2018.

\bibitem{rovio}
M.~Bloesch, M.~Burri, S.~Omari, M.~Hutter, and R.~Siegwart, ``Iterated extended
  {Kalman} filter based visual-inertial odometry using direct photometric
  feedback,'' \emph{Int. J. Robot. Res.}, vol.~36, 2017.

\bibitem{rebivo}
J.~J. Tarrio and S.~Pedre, ``Realtime edge based visual inertial odometry for
  {MAV} teleoperation in indoor environments,'' \emph{J. Intell. Robot. Syst.},
  pp. 235--252, 2017.

\bibitem{Delmerico:254865}
J.~Delmerico and D.~Scaramuzza, ``A benchmark comparison of monocular
  visual-inertial odometry algorithms for flying robots,'' in \emph{Proc.
  ICRA}, 2018.

\bibitem{ShkurtiCRV2011}
F.~Shkurti, I.~Rekleitis, and G.~Dudek, ``Feature tracking evaluation for pose
  estimation in underwater environments,'' in \emph{Canadian Conference on
  Computer and Robot Vision (CRV)}, St. John, NF Canada, 2011, pp. 160--167.

\bibitem{QuattriniLiOceans2016}
A.~{Quattrini Li}, A.~Coskun, S.~M. Doherty, S.~Ghasemlou, A.~S. Jagtap,
  M.~Modasshir, S.~Rahman, A.~Singh, M.~Xanthidis, J.~M. O’Kane, and
  I.~Rekleitis, ``Vision-based shipwreck mapping: on evaluating features
  quality and open source state estimation packages,'' in \emph{MTS/IEEE OCEANS
  - Monterrey}, Sep. 2016, pp. 1--10.

\bibitem{cummins2008fab}
M.~Cummins and P.~Newman, ``{FAB-MAP}: Probabilistic localization and mapping
  in the space of appearance,'' \emph{Int. J. Robot. Res.}, vol.~27, no.~6, pp.
  647--665, 2008.

\bibitem{cummins2011appearance}
------, ``Appearance-only {SLAM} at large scale with {FAB-MAP} 2.0,''
  \emph{Int. J. Robot. Res.}, vol.~30, no.~9, pp. 1100--1123, 2011.

\bibitem{pizer1987adaptive}
S.~M. Pizer, E.~P. Amburn, J.~D. Austin, R.~Cromartie, A.~Geselowitz, T.~Greer,
  B.~ter Haar~Romeny, J.~B. Zimmerman, and K.~Zuiderveld, ``Adaptive histogram
  equalization and its variations,'' \emph{Computer vision, graphics, and image
  processing}, vol.~39, no.~3, pp. 355--368, 1987.

\bibitem{forster2017manifold}
C.~Forster, L.~Carlone, F.~Dellaert, and D.~Scaramuzza, ``On-manifold
  preintegration for real-time visual--inertial odometry,'' \emph{{IEEE} Trans.
  Robot.}, vol.~33, no.~1, pp. 1--21, 2017.

\bibitem{ceres}
S.~Agarwal, K.~Mierle, and Others, ``{Ceres Solver},''
  \url{http://ceres-solver.org}, 2015.

\bibitem{galvez2012bagofwords}
D.~G{\'a}lvez-L{\'o}pez and J.~D. Tardos, ``Bags of binary words for fast place
  recognition in image sequences,'' \emph{{IEEE} Trans. Robot.}, vol.~28,
  no.~5, pp. 1188--1197, 2012.

\bibitem{strasdat2012local}
H.~Strasdat, ``Local accuracy and global consistency for efficient visual
  slam,'' Ph.D. dissertation, Citeseer, 2012.

\bibitem{msckf-danilidis}
{Research group of Prof. Kostas Daniilidis}, ``{Monocular MSCKF ROS node},''
  \url{https://github.com/daniilidis-group/msckf_mono}, 2018.

\bibitem{Umeyama:1991:LET:105514.105525}
S.~Umeyama, ``Least-squares estimation of transformation parameters between two
  point patterns,'' \emph{{IEEE} Trans. Pattern Anal. Mach. Intell.}, vol.~13,
  no.~4, pp. 376--380, 1991.

\bibitem{RekleitisIROS2009b}
J.~Sattar, G.~Dudek, O.~Chiu, I.~Rekleitis, P.~Giguere, A.~Mills, N.~Plamondon,
  C.~Prahacs, Y.~Girdhar, M.~Nahon, and J.-P. Lobos, ``Enabling autonomous
  capabilities in underwater robotics,'' in \emph{Proc. IROS}, 2008, pp.
  3628--3634.

\end{thebibliography}

\end{document}